\documentclass[journal]{IEEEtran}

\usepackage{graphicx}
\usepackage{float}
\usepackage[caption=false]{subfig}
\usepackage{amsmath}
\usepackage{amssymb}
\usepackage{booktabs}
\usepackage{diagbox}
\usepackage{comment}
\usepackage{multirow}
\usepackage{makecell}
\usepackage{tabularx}
\usepackage{makecell}
\usepackage{ulem}
\useunder{\uline}{\ul}{}
\usepackage[dvipsnames]{xcolor}

\definecolor{detcolor}{RGB}{183, 148, 145}%{229, 185, 181}
\definecolor{desccolor}{RGB}{147,169,102}
\definecolor{thencolor}{RGB}{0, 0, 255}



\newcommand{\gary}[1]{\textcolor{black}{#1}}
\newcommand{\qy}[1]{\textcolor{black}{#1}}

\newcommand{\revise}[1]{\textcolor{black}{#1}}

\newcommand{\new}[1]{\textcolor{black}{#1}}
\newcommand{\nickname}{Box2Seg}

\usepackage[capitalize]{cleveref}
\crefname{section}{Sec.}{Secs.}
\Crefname{section}{Section}{Sections}
\Crefname{table}{Table}{Tables}
\crefname{table}{Tab.}{Tabs.}

\begin{document}

\title{Box2Seg: Learning Semantics of 3D Point Clouds with Box-Level Supervision}

\author{Yan~Liu$^{1}$, 
        Qingyong~Hu$^{2}$,
        Yinjie~Lei$^3$, 
        Kai~Xu$^4$, 
        Jonathan~Li$^5$,
        Yulan~Guo$^{1,4,\dagger}$\\
        $^1$Sun Yat-Sen University
        \quad 
        $^2$University of Oxford
        \quad 
        $^3$Sichuan University
        \quad 
        $^4$National University of Defense Technology
        \quad 
        $^5$University of Waterloo 
        % <-this % stops a space
        
        \thanks{$\dagger$ denotes the corresponding author.}
}

\maketitle

%%%%%%%%% ABSTRACT
\begin{abstract}
Learning dense point-wise semantics from unstructured 3D point clouds with fewer labels, although a realistic problem, has been under-explored in literature. While existing weakly supervised methods can effectively learn semantics with only a small fraction of point-level annotations, we find that the vanilla bounding box-level annotation is also informative for semantic segmentation of large-scale 3D point clouds. In this paper, we introduce a neural architecture, termed Box2Seg, to learn point-level semantics of 3D point clouds with bounding box-level supervision. The key to our approach is to generate accurate pseudo labels by exploring the geometric and topological structure inside and outside each bounding box. Specifically, an attention-based self-training (AST) technique and Point Class Activation Mapping (PCAM) are utilized to estimate pseudo-labels. The network is further trained and refined with pseudo labels. Experiments on two large-scale benchmarks including S3DIS and ScanNet demonstrate the competitive performance of the proposed method. In particular, the proposed network can be trained with cheap, or even off-the-shelf bounding box-level annotations and subcloud-level tags.
\end{abstract}

%%%%%%%%% BODY TEXT
\section{Introduction}
\gary{Semantic understanding of the 3D environment is a key enabler for robots to perceive and interact with the physical world. A number of real applications such as robotic grasping \cite{grasping}, autonomous navigation \cite{navigation}, and human-machine interaction \cite{human_machine} require the machine to precisely recognize (\textit{i.e.}, dense segmentation) its 3D surroundings. However, this remains challenging due to the complex geometrical structures of environments and limited capacity of existing 3D understanding models.}

\gary{As 3D data acquisition and annotation become increasingly affordable, remarkable progress has been achieved in the task of 3D point cloud analysis in recent years \cite{guo2019deep}. This is not only reflected in the emergence of several milestone network architectures \cite{qi2017pointnet, qi2017pointnet++, li2018pointcnn, wu2018pointconv, thomas2019kpconv, wu2018squeezeseg, sparse, 4dMinkpwski, Point_voxel_cnn}, but also a number of high-quality densely annotated public datasets \cite{Semantic3D, behley2019semantickitti, hu2020towards, Dai2017scannet, 2D-3D-S, chang2015shapenet, Waymo, caesar2020nuscenes, mo2019partnet}. However, as suggested in \cite{hu2020towards}, it is still an open question that if existing 3D segmentation pipelines can be generalized/scaled to extremely large-scale real 3D environments with complex geometrical structures. One main issue is that most existing pipelines still heavily rely on the strong supervision signals (\textit{i.e.}, point-wise dense semantic labels), which are not always available and easy to collect in practice. Furthermore, the requirement of strong supervision also prevents existing approaches from taking full advantage of different forms of rich annotations in existing datasets. For example, there are more than 12 million bounding boxes in the Waymo dataset \cite{Waymo}, while not fully exploited for segmentation tasks.}

\begin{figure}[t]
\centering
\includegraphics[width=1.0\linewidth]{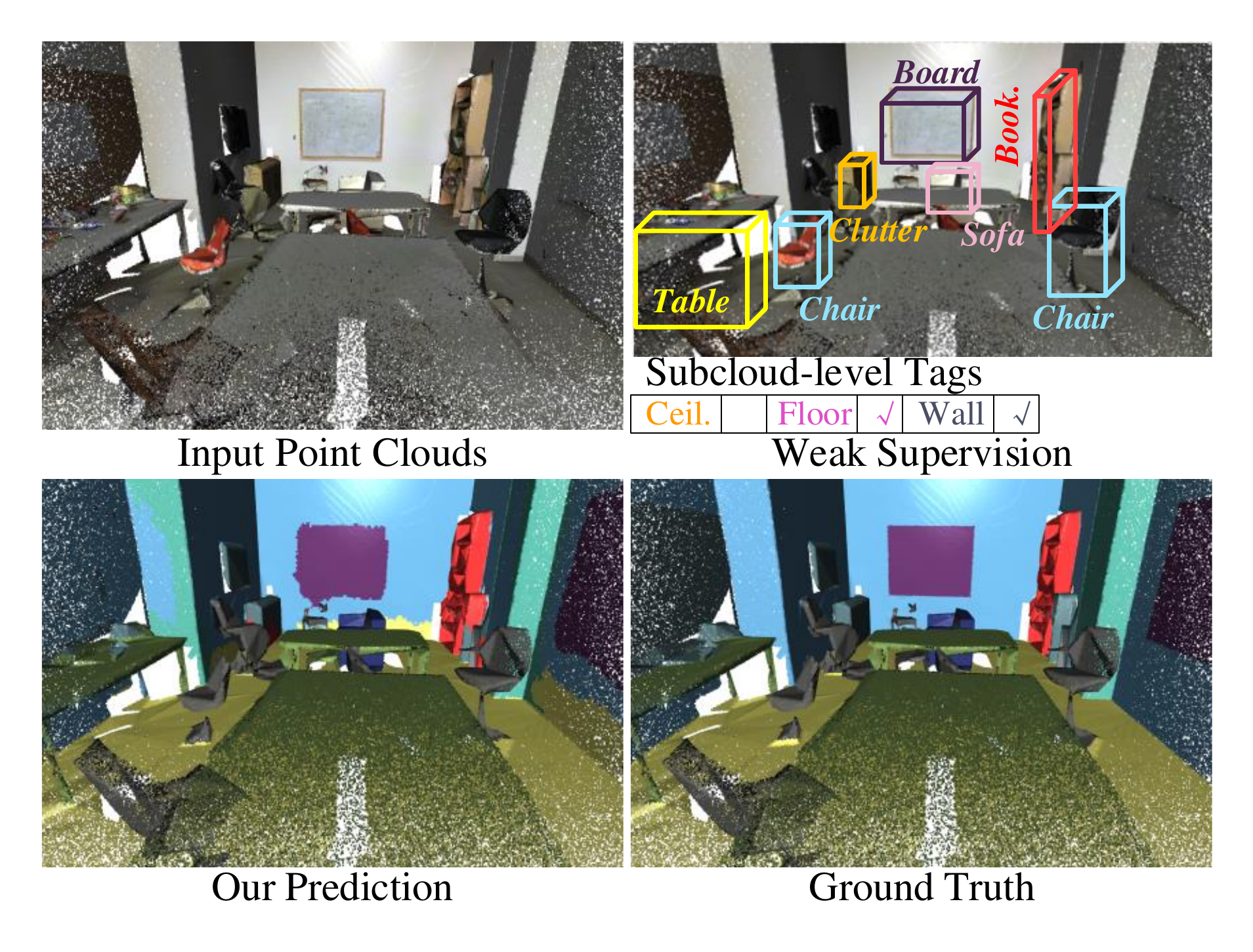}
\caption{\new{An illustration of the weak supervision setting of our \nickname{}. Unlike full annotation strategies that require point-wise semantic labels, our \nickname{} can be trained with low-cost bounding box annotations and subcloud-level tags, but can also achieve reasonable semantic segmentation performance.} \label{fig:label-strategy}}
\end{figure}

\gary{A handful of recent works have started to explore the weakly supervised semantic segmentation of 3D point clouds \cite{hu2021sqn, xu2020weakly, hou2020exploring, tao2020seggroup, spatial_layout, zhang2021perturbed, zhang2021weakly, cheng2021sspc, liu2021one}. That is, learning semantics from other indirect formats of weak supervision or a small fraction of labeled points. MPRM \cite{mprm} and SegGroup \cite{tao2020seggroup} are proposed to generate pseudo point-level labels from subcloud-level labels or seg-level labels. SQN \cite{hu2021sqn}, Xu's work\cite{xu2020weakly} and SceneContrast\cite{hou2020exploring} are introduced to learn semantic segmentation from a small fraction of randomly annotated or actively labeled points. This is achieved by leveraging the semantic similarity within a neighborhood or gradient approximation, and contrastive pretraining followed with finetuning. Despite the promising results, it is still non-trivial to deploy these weak supervision schemes in practice as a variety of customized tools are unavailable yet (\textit{e.g.} It remains challenging for existing tools to annotate sparsely distributed points or subcloud labels on-the-fly). Different from the aforementioned scheme, we instead turn our eyes to the commonly-used bounding boxes in 3D object detection. \textit{Is it possible to learn point-level semantics from the bounding box annotations?}}

\new{In this paper, we propose a weakly supervised semantic segmentation framework called \nickname{} by leveraging the \textit{commonly used} yet \textit{low-cost} bounding box annotations and subcloud-level tags as the supervision signal. Overall, the key idea is to fully exploit the geometric priors and structural cues from the limited supervision signals, further estimate the pseudo labels of the foreground points and background points separately in a divide-and-conquer manner. %\ly{We first introduce a 3D Grabcut method to separate points inside bounding boxes to capture relatively accurate pseudo labels for model training, which shows the potential of leveraging box-level supervision. On this basis, we further explore a self-training-based approach to generate point-wise pseudo-labels with pseudo labeling and attention mechanism. These techniques are effective in eliminating the negative effects caused by points with ambiguous even incorrect pseudo labels.} 
% In particular, we first explore two methods for foreground pseudo label generation, including an unsupervised 3D Grabcut and a learnable attention-based self-training pipeline. These techniques are effective in separating the foreground points from the backgrounds, and reducing the ambiguity of points that lies in overlapped regions of multiple bounding boxes.
% Moreover, we also train a dedicated point classifier to mine the localization cues by investigating the generated point class activation maps, hence further estimate the pseudo labels of background points. Finally, the refined generated point-wise pseudo labels allow us to train a segmentation network in a fully supervised way, although the supervision signal is not perfectly correct. Experiments on several large-scale open benchmarks demonstrate the feasibility of our solution. The proposed \nickname{} framework can even achieve comparable performance with several early supervised baseline networks on existing datasets, by learning from limited supervision.
In particular, we first explore two methods for foreground pseudo label generation, including an unsupervised 3D Grabcut and a learnable attention-based self-training pipeline. We then train a dedicated point classifier to mine the localization cues by investigating the generated point class activation maps, hence further estimate the pseudo labels of background points. Finally, the refined generated point-wise pseudo labels allow us to train a segmentation network in a fully supervised way. Experiments on several large-scale open benchmarks demonstrate the feasibility of our solution. The proposed \nickname{} framework can even achieve comparable performance with several early supervised baseline networks on existing datasets, by learning from limited supervision as shown in Fig. \ref{fig:label-strategy}.
}

\gary{Overall, the contributions of our paper can be summarized as follows:}

\begin{itemize}
\setlength{\itemsep}{0pt}
\setlength{\parsep}{0pt}
\setlength{\parskip}{0pt}
    \item \gary{The proposed \nickname{} framework is the first attempt to achieve semantic segmentation of large-scale 3D point clouds, with the usage of only bounding box-level annotations and subcloud-level tags.}

    \item \new{An effective approach is proposed to generate pseudo labels for both foreground and background points with Attention-based Self-Training (AST) and Point Class Activation Maps (PCAMs).}
    
    \item \gary{Extensive experiments conducted on two large-scale point cloud datasets and various backbones demonstrate the effectiveness and versatility of \nickname{}. In particular, the proposed framework allows leveraging the ready-to-use bounding box labels in existing large-scale 3D object detection datasets.}

\end{itemize}

%------------------------------------------------------------------------
\section{Related Works}
We retrospect recent mainstream semantic segmentation approaches, which can be categorized into two types according to supervision (full supervision or weak supervision).

%\textbf{Semantic Segmentation of 3D Point Clouds with Full Supervision.}
\subsection{Semantic Segmentation with Full Supervision}

\gary{Recently, pioneering works such as PointNet \cite{qi2017pointnet} and SparseConvNet \cite{sparse} have greatly facilitated the development of deep learning in semantic segmentation of 3D point clouds. A number of sophisticated approaches \cite{qi2017pointnet++, wu2018pointconv, li2018pointcnn, thomas2019kpconv, hu2019randla, 4dMinkpwski, Point_voxel_cnn, cylinder3d} have been further proposed to substantially improve the runtime performance from various aspects. Following \cite{guo2019deep}, existing approaches can be roughly divided into four categories according to the representation of 3D point clouds used in their frameworks: 1) point-based methods \cite{qi2017pointnet, qi2017pointnet++, li2018pointcnn, zhao2019pointweb, wu2018pointconv, hu2019randla, thomas2019kpconv}, 2) voxel-based methods \cite{sparse, vvnet, 4dMinkpwski, han2020occuseg}, 3) projection-based methods \cite{wu2019squeezesegv2, xu2020squeezesegv3, boulch2017unstructured}, and 4) hybrid methods \cite{Point_voxel_cnn, e3d}.}

\gary{Although remarkable progress has been achieved, these methods usually require point-level semantic labels to provide dense supervision signal for network training. This supervision signal is usually extremely expensive and not always available in practice. This issue is particularly serious for customized practical applications since not everyone has a dedicated annotation team. Motivated by this, we propose \nickname{}, a weakly-supervised framework to learn from the ready-to-use bounding box annotations and cheap subcloud-level tags.}

\begin{figure*}
\centering
\includegraphics[width=1.0\linewidth]{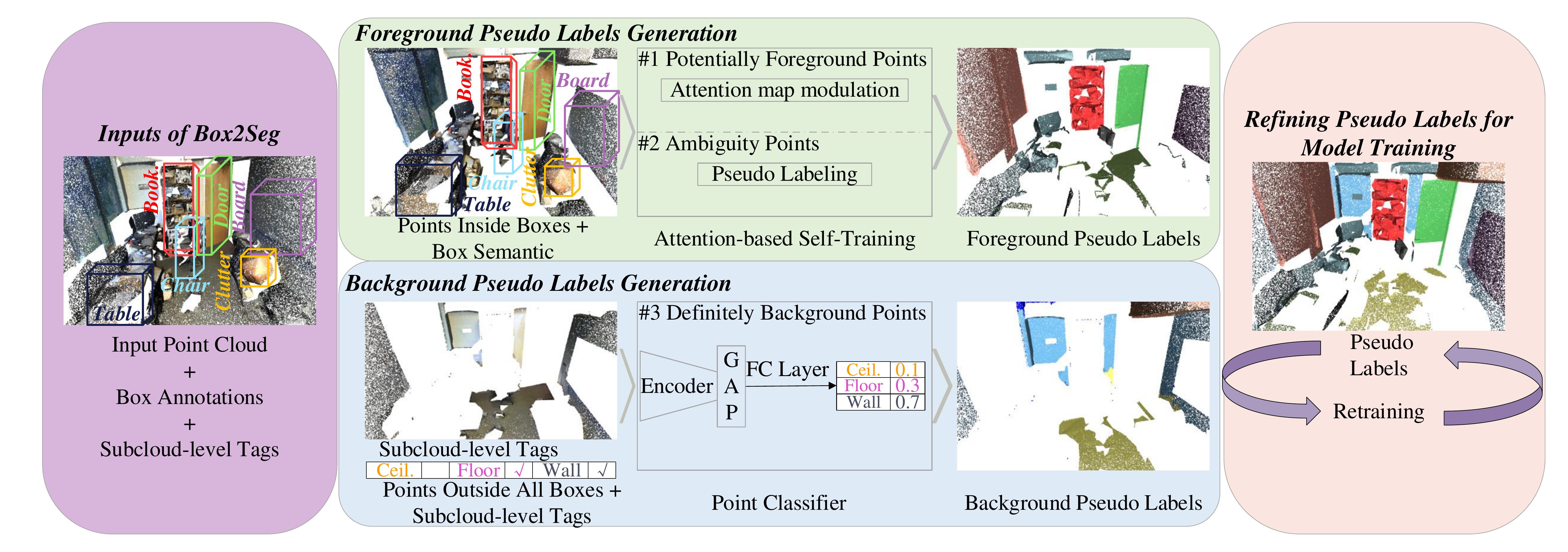}
\caption{\gary{Overall framework of the proposed \nickname{}. Unlike conventional fully-supervised point cloud semantic segmentation methods that use point-level annotation for model training, our \nickname{} takes the bounding-box annotations and subcloud-level tags as its main supervision signals. The whole pipeline consists of three steps, including learning foreground pseudo labels from box-level annotations (Sec.\ref{sec:fore-ground-generation}), learning background pseudo labels  from  subcloud-level tags (Sec.\ref{sec:background-pseudo-generation}), and refining pseudo labels for model training (Sec.\ref{sec:refine-and-training}).
%foreground pseudo label generation, background pseudo label generation, and the re-training of segmentation models. 
Attention mechanism, pseudo labeling, and point class activation map techniques are adopted.
}}
\label{fig:pileline}
\end{figure*}

\subsection{Semantic Segmentation with Weak Supervision}
%\textbf{\qy{Semantic Segmentation of 3D Point Clouds with Weak Supervision.}}
\gary{ To reduce the annotation cost, a handful of recent methods \cite{xu2020weakly, hu2021sqn, hou2020exploring, tao2020seggroup, mprm, spatial_layout} have been proposed to investigate weakly supervised semantic segmentation of 3D point clouds. These existing approaches can be divided into two categories according to their weak supervision settings: methods with limited point supervision, methods with inexact supervision.}

\gary{\textbf{Methods with Limited Point Supervision.} These methods aim at learning the semantics of 3D point clouds using a small fraction of labeled points (\textit{e.g.} 10\%, 1\textperthousand{}) as their supervision. In general, the labeled points can be determined randomly \cite{hu2021sqn, xu2020weakly} or through active learning techniques \cite{hou2020exploring}. Xu et al. \cite{xu2020weakly} achieve weakly supervised point cloud segmentation with 10$\times$ fewer labels through gradient approximation and spatial color consistency. Hu et al. \cite{hu2021sqn} empirically identify the redundancy in dense 3D annotations, and propose a network (namely, SQN) to implicitly augment the supervision signal by leveraging the semantic similarity within neighboring points. Inspired by the unsupervised PointContrast \cite{xie2020pointcontrast} framework, Hou et al. \cite{hou2020exploring} further investigate label-efficient learning on 3D point clouds and active labeling techniques.}

\gary{\textbf{Methods with Inexact Supervision.} Instead of using point-level supervision, these methods attempt to learn from various inexact supervision, including subcloud-level tags \cite{spatial_layout,mprm}, seg-level labels \cite{tao2020seggroup}, and 2D pixel-wise labels \cite{wang2020weakly}. Wei et al. \cite{wei2020multipath} first train a classifier using subcloud-level labels, and then introduce a multi-path region mining module to extract the object localization region cues. They finally generate pseudo labels for the training of their segmentation network. Ren et al. \cite{spatial_layout} introduce WyPR to achieve semantic segmentation, 3D proposal generation, and 3D object detection jointly, with only subcloud-level tags being provided. Tao et al \cite{tao2020seggroup} propose SegGroup to achieve 3D semantic segmentation and instance segmentation, by taking oversegmentation as the pre-processor and seg-level labels as the supervision signals. Wang et al. \cite{wang2020weakly} propose a Graph-based Pyramid Feature Network (GPFN) to learn semantics from 3D point clouds with solely 2D supervision. } 

\revise{Unlike existing techniques, the proposed Box2Seg is the first attempt to explore the weak-supervision signals composed of bounding box annotation and subcloud-level tags in the field of 3D point cloud segmentation. Firstly, we attempt to leverage box supervision to tackle raw points directly. Besides, a conceptually simple yet effective framework based on PCAM is also presented. We combine 3D GrabCut and PCAM as the baseline method to obtain a segmentation model. Further, we introduce  pseudo labels and attention mechanism to regularize the training phase. In particular, this paper aims at building a weakly supervised framework to learn semantics from cheap bounding box labels and subcloud-level tags, in light of the large quantities of ready-to-use bounding box annotations (\textit{e.g.} 12 million bounding boxes in the Waymo dataset \cite{Waymo}) in existing 3D object detection datasets \cite{caesar2020nuscenes, Waymo}.} 

% We hope our Box2Seg can be regard as a starting point, stimulating further research works in this topic, eventually bridged the gap between the tasks of 3D object detection and dense point cloud segmentation.}

%\gary{Analogous to these methods with inexact supervision, we introduce \nickname{} in this paper to achieve semantic segmentation of 3D point clouds, by leveraging annotated instance bounding boxes and scene level tags as its supervision. }

\section{The Proposed Method}

\subsection{\qy{Overview}}

\new{The pipeline of our \nickname{} is depicted in Figure \ref{fig:pileline}. The main idea of our approach is to generate point-wise pseudo labels inside and outside each bounding box in a divide-and-conquer manner, by fully exploring the geometric priors and local topological structures in point clouds. Specifically, given a 3D point cloud $\mathcal{P}$ and its associated bounding box annotations (\textit{i.e.}, 3D coordinates of the bounding box and corresponding semantics), we first divide all points into three categories according to the layout and spatial relationship of the bounding box in the 3D space: 1) \textbf{Potentially foreground points}. A point will be classified into this category if and only if this point is located within a unique bounding box. 2) \textbf{Ambiguity points}. These points are usually located at the overlapped area of multiple bounding boxes, due to the inaccurate bounding box annotations or possible outliers. 3) \textbf{Definitely background points}. We assume a point belongs to this category if it does not fall within any bounding box. Up to here, the next step is to estimate pseudo labels for each category of points.}

%\gary{The pipeline of our \nickname{} is depicted in Figure \ref{fig:pileline}. It is noted that the key to our approach is to generate point-wise pseudo labels inside and outside each bounding box in a divide-and-conquer manner, by fully exploring the geometric priors and local topological structures in point clouds. \ly{In particular, we first adopt 3D Grabcut \cite{rother2004grabcut} to verify the possibility of using box-level supervision in semantic segmentation. We further propose a self-training-based method.} Then, we leverage the point class activation maps to further estimate the background pseudo labels. Finally, these per-point pseudo labels are further combined to train an existing segmentation network.}

% \todo{Existing point cloud annotation tools like 3D BAT \cite{zimmer20193d} can label 3D bounding boxes for point cloud detection. Besides, this software provides 3D bounding box annotation cost, for an experienced worker, labeling 388 objects in the Nuscenes dataset \cite{} spends about 49 minutes (7.57s per instance). It means harnessing 3D bounding boxes to annotate a signal scene of Scannet costs about 3.2 minutes.}. \textbf{}

\begin{figure}[t]
\centering
\includegraphics[width=1.0\linewidth]{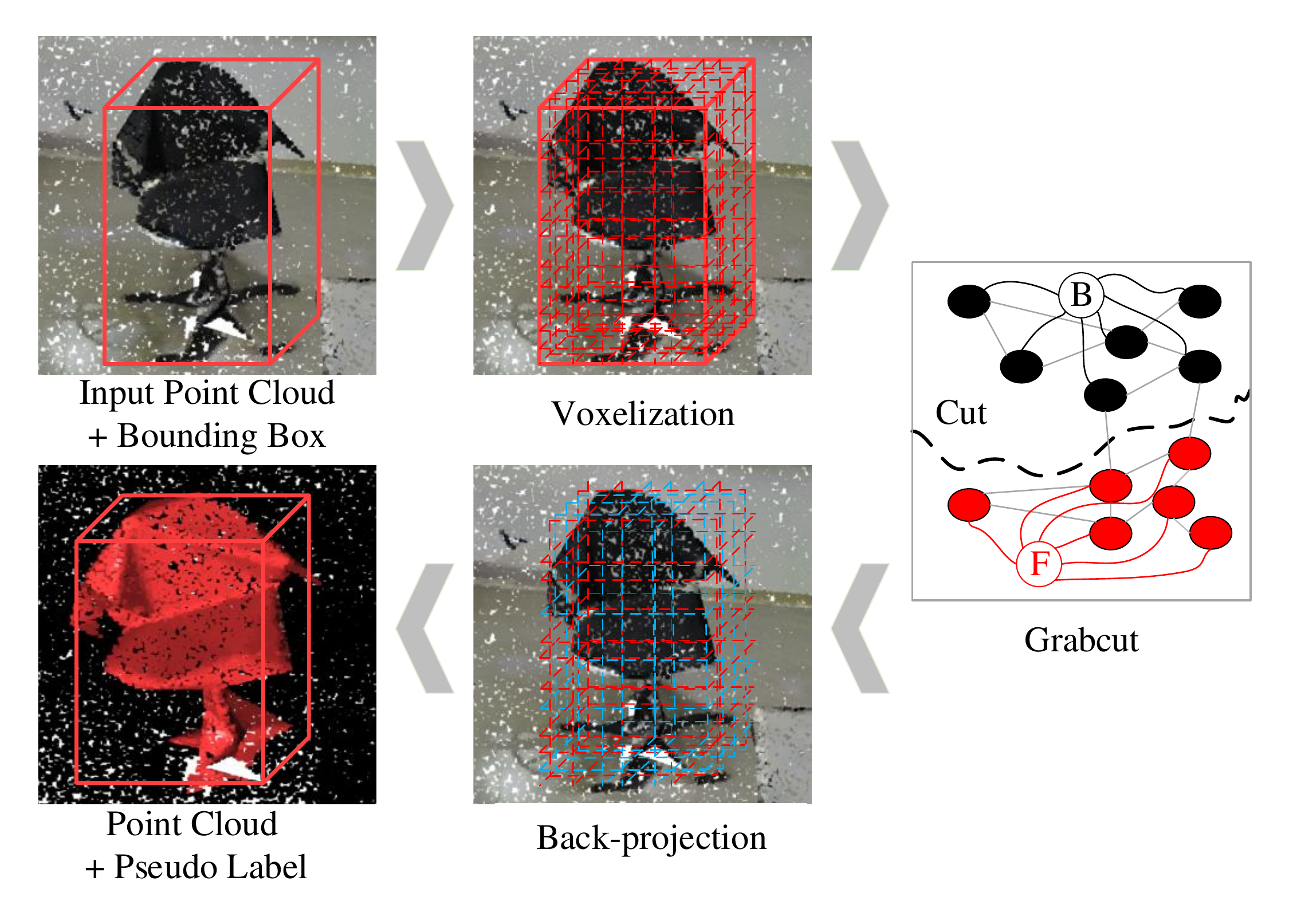}
\caption{The pipeline of the unsupervised 3D GrabCut. \label{fig:foreground-segmentation}}
\end{figure}

\subsection{Foreground Pseudo Labels Generation \label{sec:fore-ground-generation}}

\new{We first consider how to generate pseudo-labels for foreground points. The naive solution is to endow the points that fall within the bounding box with the same semantic label as the box. However, the foreground labels may be incorrectly assigned to background points in this case, since the bounding box annotations are usually not precise as expected. To this end, we explore two different solutions in our framework, including the unsupervised 3D Grabcut and a dedicated attention-based self-training pipeline. The detailed descriptions are as follows:
}

\textbf{Unsupervised 3D Grabcut.} \new{Grabcut\cite{graphcut} is an unsupervised segmentation approach, which can estimate the instance boundaries and find the optimal smooth seam for the separation of foreground and background points within a specific 2D region. Here, we extend this technique to 3D point clouds.} 

Given raw points and a number of bounding boxes with semantic labels, we aim to separate the foreground and background points within each individual bounding box. To this end, we first voxelize the point cloud within a specific bounding box, in light of the unstructured and orderless nature of 3D points. Then, each voxelized point cloud is first oversegmented into geometrically homogeneous partitions (\textit{i.e.}, superpoints) based on the traditional Simple Linear Iterative Clustering (SLIC) algorithm \cite{achanta2012slic}. Next, a graph $\mathcal{G} = (\mathcal{V},\mathcal{E})$ with nodes associated with the superpoints partitions are built, the foreground voxels are further segmented. Finally, semantic labels of these voxels are back-projected to the raw point cloud. By iteratively traversing all bounding boxes, we can obtain the point-level semantic mask for foreground points in each instance as shown in Fig. \ref{fig:foreground-segmentation}. Note that, 3D Grabcut only takes the low-level handcrafted features of the 3D point clouds, it may fail when there is no clear boundary of the foreground points and background points. Additionally, it is non-trivial to apply 3D Grabcut on the ambiguity points. In this regard, we only take the 3D Grabcut as the baseline method in our framework, and further introduce an attention-based self-training pipeline to achieve better foreground segmentation in 3D point clouds.
%Here, low-level information of points is used in 3D Grabcut, and it is possible to train a segmentation model with box supervision. To capture accurate semantics, we introduce a self-training-based method to enhance feature representation.  

\begin{figure}[t]
\centering
\includegraphics[width=1.0\linewidth]{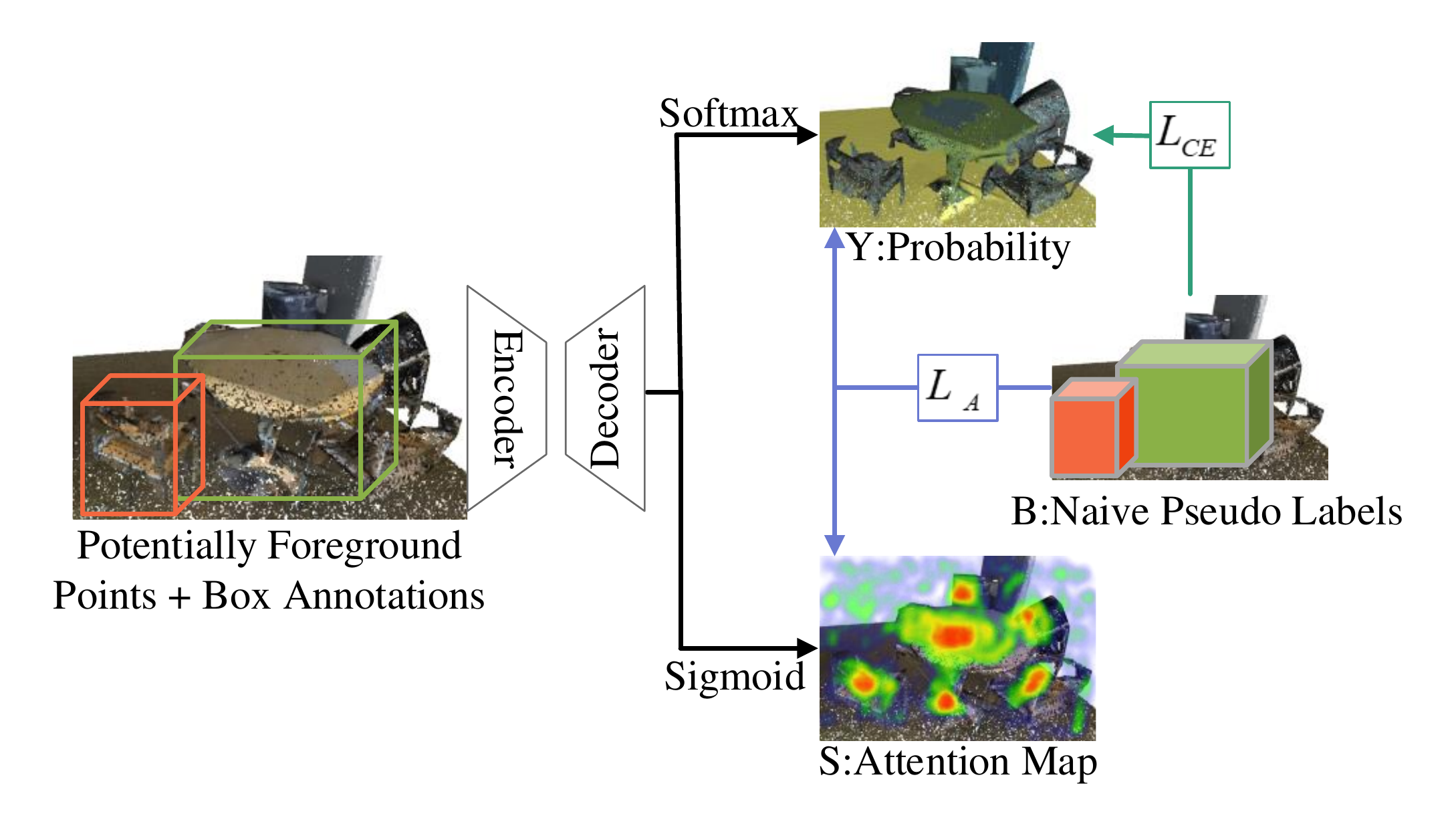}
\caption{The pipeline of the attention map modulation to separate foreground and background points inside boxes. \label{fig:attention-foreground}}
\end{figure}

%\subsubsection{Accurate Foreground Pseudo Label Generation}
%\subsubsection{Extracting Foreground Semantics for Pseudo Label Generation}

\textbf{\underline{A}ttention-based \underline{S}elf-\underline{T}raining Pipeline (AST).} \new{To precisely separate the foreground and background points within a bounding box, we propose a learnable, attention-based self-training pipeline here. This is based on the fact that the 3D instance in the bounding box usually has clear shapes, albeit far from perfect. Therefore, there should be a higher correlation between points belonging to the same instance. Motivated by this, we introduce an attention-map-based regularization scheme to modulate the learning process. Specifically, we explicitly estimate an attention map for the input point clouds, where the sigmoid function is used to output the final logits. The attention map is further used to modulate the standard cross-entropy loss. The ultimate goal is to learn feature representations similar to the attention map with clear boundaries between foreground and background points. As shown in Fig. \ref{fig:attention-foreground}, a point-wise cross-entropy loss and the predicted attention map is used for training, that is:}
\begin{align}
{L}_{A} =  - \sum_{c} S_{c}(p_{p},\theta)*{\rm log}(Y_{c}(p_{p},\theta))*B_{c},
\end{align}
\new{where $p_{p}$ is the point that located in a unique box, $S_{c}$ is the attention map for class $c$ generated by the sigmoid activation function, $Y_{c}$ is the predicted probabilities, and $B\in \{0,1\}^{C}$ is a binary mask which indicates the semantics of the bounding box (naive pseudo label), also $\theta$ are learnable parameters of segmentation model. } 

% \ly{As aforementioned, using bounding box supervision can capture coarse semantic information of foreground instances, which demonstrates the feasibility of leveraging such weak supervision. In spite of this, due to the limited supervision provided by box-level annotations, two  issues remain unsolved: 
% 1)  points covered by multiple bounding boxes have semantic ambiguity,
%  2) background points inside a bounding box are assigned with incorrect semantics. 
% To tackle these mentioned issues, we propose to leverage a self-training-based pipeline to regularize the training process.}  

\begin{figure}[t]
\centering
\includegraphics[width=1.0\linewidth]{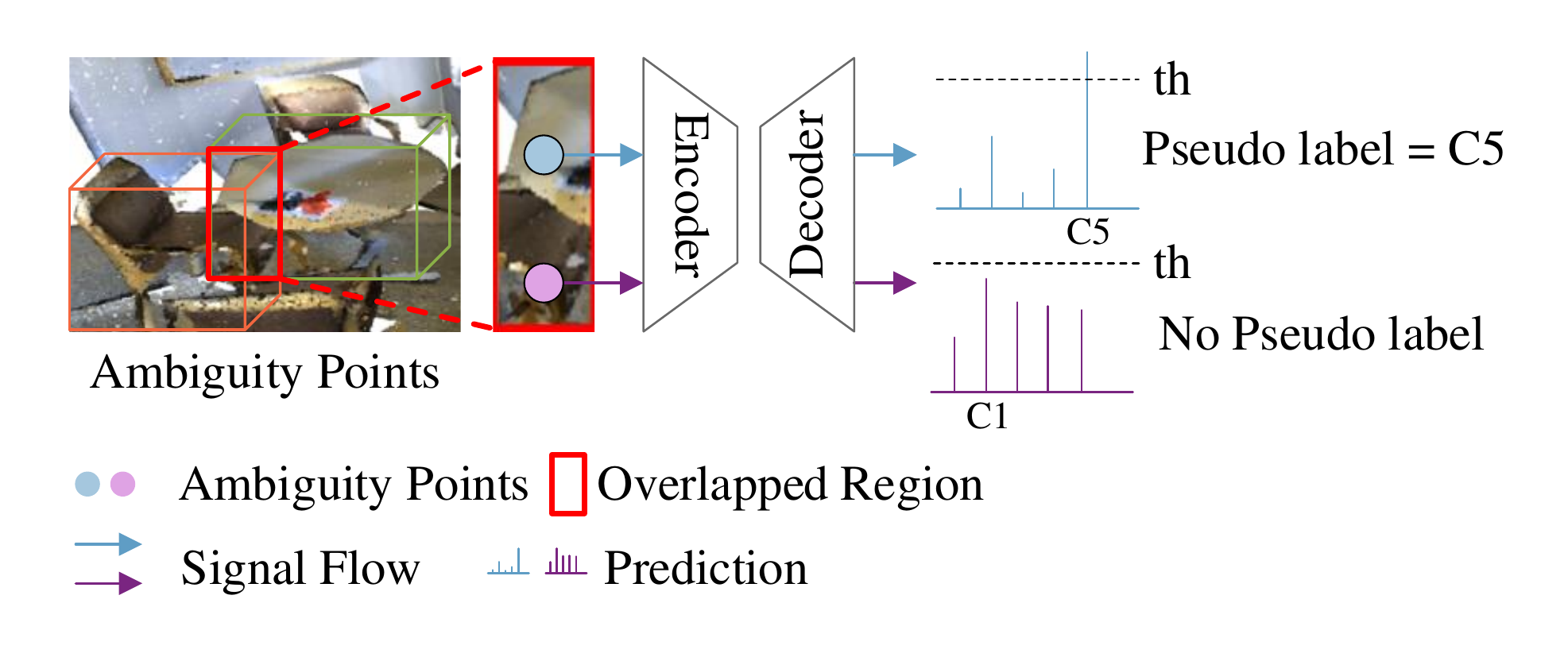}
\caption{The pipeline of the pseudo labeling method to produce pseudo labels for ambiguity points located in overlapped regions. \label{fig:overlap-pseudo-labeling}}
\end{figure}

\new{For ambiguity points lies in the overlapped region of multiple bounding boxes, we further introduce a self-training pipeline to progressively reduce the ambiguity in semantics as shown in Fig. \ref{fig:overlap-pseudo-labeling}. Inspired by semi-supervised learning frameworks \cite{MT,laine2016temporal} which train labeled and unlabeled samples together, we follow the concept of entropy minimization, to assign pseudo labels to these points according to the prediction in the further training stage. The category with the maximum predicted probability is used to determine the pseudo label of an ambiguous point, that is:}

%\ly{\textbf{- Pseudo Labeling: Reducing ambiguity in overlapped points.} We assign points located in a single bounding box with the  semantic label of the box. However, the semantic labels of these points surrounded by more than one boxes cannot be uniquely assigned temporarily. Inspired by semi-supervised learning frameworks \cite{MT,laine2016temporal} train labeled and unlabeled samples together, we adopt entropy minimization to assign pseudo labels to these points according to the prediction in the further training stage. The category with the maximum predicted probability is used to determine the pseudo label of an ambiguous point, that is:
\begin{align}
\widetilde{y}_{p_{o}} =   \underset{c}{\mathrm{argmax} } (Y_{c}(p_{o},\theta)),
\end{align}
where $p_{o}$ is a point in the overlapped region, and $\widetilde{y}_{p_{o}}$ is the generated pseudo label. Note that, incorrect prediction can mislead the training phase in the early stage. To avoid this problem, we only allow points with relatively high confidence values (\textit{i.e.}, maximum likelihood larger than a predefined threshold of 0.8) for training. If the confidence of a point is low, no pseudo label is assigned to this point in the training process.

\subsection{\qy{ Background Pseudo Labels Generation \label{sec:background-pseudo-generation}}}

\new{Apart from these foreground points, a raw point cloud also contains a large number of background points (\textit{e.g.} \textit{floor}, \textit{ceiling}, \textit{etc.}). In general, the background is usually composed of amorphous areas of homogenous or repeating geometric patterns. Due to unclear borders and irregular shapes, there are usually no bounding box annotations for these categories. As a workaround, we propose to learn the semantics of background from subcloud-level class tags $\mathbf{y} _{c}\in \left \{ 0,1 \right \} ^{C}$, which indicates whether a specific class appears in the subcloud or not. Inspired by the success of Class Activation Mapping (CAM) techniques \cite{zhou2016learning} in weakly supervised semantic segmentation of 2D images, we first train a point classifier with the subcloud-level supervision to estimate whether specific background categories appear. Then, we further exploit the localization cues of specific classes from the weighted combination of intermediate latent feature maps of the trained classifier. Here, the cross-entropy loss is used under the multi-class classification setting.}

\begin{figure}[t]
\includegraphics[width=0.9\linewidth]{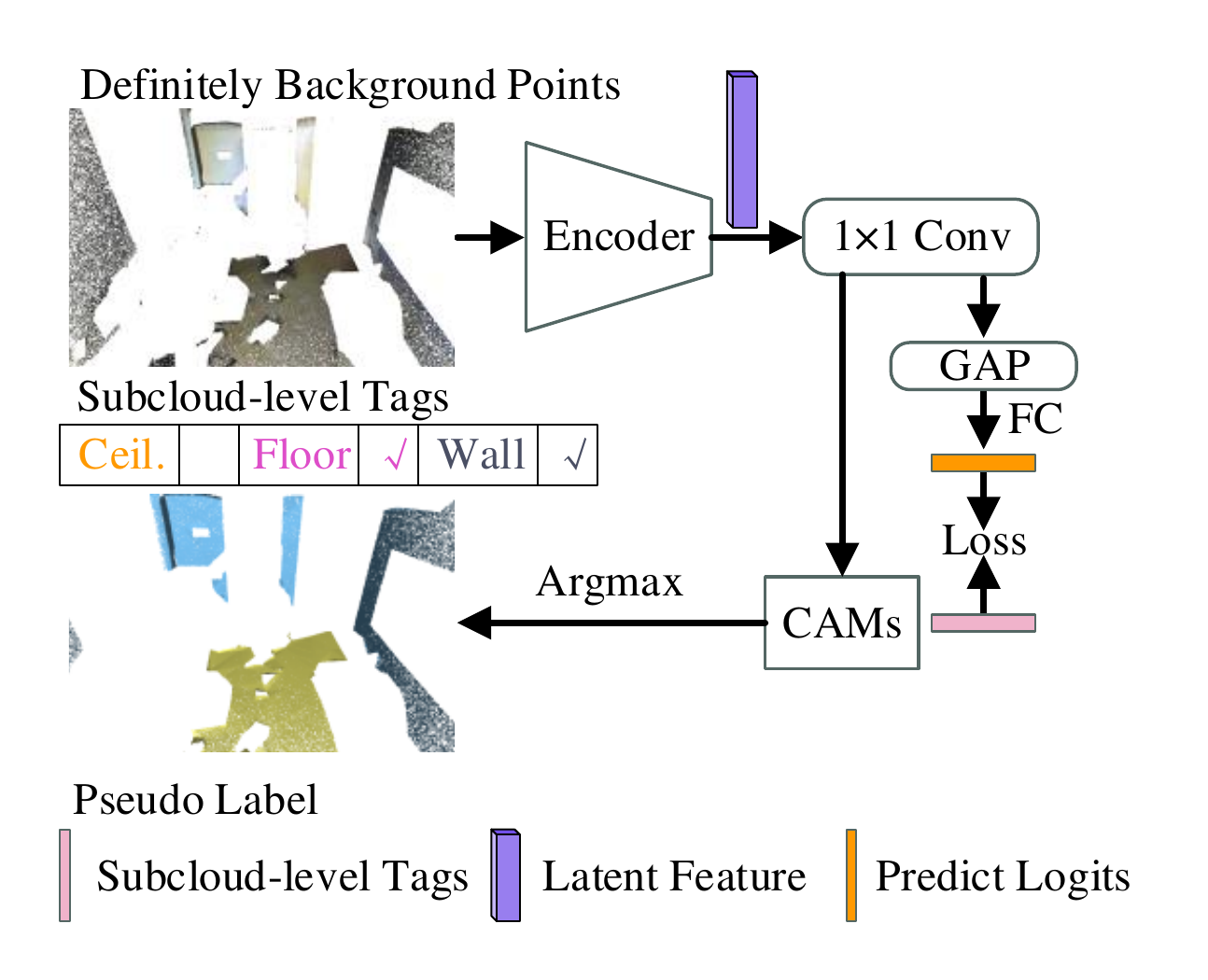}
\caption{\gary{The background pseudo label generation pipeline. Note that, this is only used to infer the semantics for points outside bounding boxes. The classifier only used for pseudo labels generation. \label{fig:background_cam_generation}}}
\end{figure}

\new{As shown in Fig. \ref{fig:background_cam_generation}, the geometrical patterns of the input point clouds $\mathcal{P}$ are first hierarchically encoded as compact latent vectors. Considering the efficiency and the capacity of processing large-scale point clouds, we adopt RandLA-Net \cite{hu2019randla} as the backbone for feature extraction. Then, instead of the symmetry decoder module used in the U-Net architecture \cite{ronneberger2015u}, we append several extra 1$\times$1 convolution layers to further predict the classification probabilities. Following \cite{zhou2016learning}, Global Average Pooling (GAP) is further utilized to aggregate feature maps, and the sigmoid cross-entropy loss is used to train the classifier. Note that, this classifier is only used for background pseudo labels generation.}

%\gary{As shown in Fig. \ref{fig:background_cam_generation}, we first hierarchically encode the geometrical patterns of point cloud $\mathcal{P}$ as compact latent vectors. Considering the efficiency and the capacity of processing large-scale point clouds, we adopt RandLA-Net \cite{hu2019randla} as the backbone of our framework. Then, instead of the symmetry decoder module used in the U-Net architecture \cite{ronneberger2015u}, we append several 1$\times$1 convolution layers to further predict the classification probabilities. Note that, Global Average Pooling (GAP) is used at the final stage, and the sigmoid cross-entropy loss is used to train the classifier. Besides, the classifier only used for background pseudo labels generation.}

\new{Once the training of the point classifier is completed, the class activation map of point $p$ with respect to semantic class $c$ is given as:}

\begin{align}
M_{c}(p_{b}) & = w_{c}^{\mathrm{T}}\cdot f_{cam} (p_{b})\cdot \mathbf{y} _{c}, 
\end{align}
\gary{where $p_{b}$ is the point outside all boxes, $M_{c}(p_{b})$ is the class activation map for class $c$ at point $p_{b}$, $w_{c}^{\mathrm{T}}$ is the parameters of the classification layer linked to class $c$, $f_{cam} (p_{b})$ is the last latent feature fed into the global average pooling layer. $\mathbf{y} _{c}$ is an one-hot vector, which indicates the presence or absence of semantic category $c$. Therefore, we can further determine the pseudo label of each point by assigning the class with the largest activation value:}

\begin{align}
\widetilde{y}_{p_{b}} =  \underset{c}{\mathrm{argmax} } (M_{c}(p_{b})).
\end{align}

\gary{Note that, the pseudo labels generated at this stage are only used to determine the semantics of background points (\textit{i.e.}, points outside bounding boxes), while the semantics of  foreground points remain unchanged (\textit{i.e.}, already determined in Sec. \ref{sec:fore-ground-generation}).}

\subsection{ Refining Pseudo Labels for Model Training \label{sec:refine-and-training}}

%In reality, ideal bounding boxes can cover instance points accurately as instances are separated in 3D space. 
\new{Considering the PCAM only captures the most discriminative regions in subclouds, we further propose a pseudo label refining module to select the most reliable predictions, since not all pseudo labels generated by PCAM are reliable. Inspired by the recent progress achieved in weakly-supervised semantic segmentation \cite{hu2021sqn}, where only a small fraction of correctly labeled points are sufficient to train a semantic segmentation model for point clouds. Therefore, we select the top 20\% points with the highest prediction confidences and use their pseudo labels for training, and empirically find the supervision signal is sufficient.}

%are not correct as we expect. As the PCAM only captures the most discriminating region in subclouds, using the generated pseudo labels directly may produce wrong information in the training stages. As mentioned in previous work \cite{xu2020weakly}, only a small fraction of points can still capture semantics efficiently. To obtain the reliable training signal, we select the top 20\% points with highest predicted confidence value to provide their pseudo labels in the training stage. }

% \subsection{\gary{Learning Semantic Segmentation from Pseudo Labels}}
\new{Once the point-wise pseudo labels of the raw point clouds are obtained, we can retrain a network with the generated pseudo labels in a fully supervised manner. Specifically, we take all points for training, but only use the generated foreground pseudo labels and selected background pseudo labels for the calculation of vanilla cross-entropy loss $L_{CE}$.} Besides, the attention mechanism is used to regularize the training phase, the overall training loss function is shown as:

\begin{align}
{L} = {L_{CE} + \alpha L_{A}} ,
\end{align}
where $\alpha$ is a weight applied to the attention loss.

{Finally, the trained model is used to infer the semantic labels of the raw point clouds. Note that, we are aware that the generated pseudo labels are not completely correct, but similar to \cite{mprm}, we found that the final segmentation results are totally acceptable, even trained with these flawed labels. }
\begin{table*}[thb]
\centering
\resizebox{0.95\textwidth}{!}{%
\begin{tabular}{c|rcccccccccccccc}
\Xhline{3.0\arrayrulewidth}
Settings & Methods & mIoU(\%)  & \textit{ceil}. & \textit{floor}  & \textit{wall}  & \textit{beam}  & \textit{col}. & \textit{win}. & \textit{door} & \textit{table} & \textit{chair} & \textit{sofa} & \textit{book}. & \textit{board} & \textit{clutter} \\ \Xhline{2.0\arrayrulewidth}
\multirow{12}{*}{\begin{tabular}[c]{@{}c@{}}Full Supervision \\ (100\%)\end{tabular}} & PointNet \cite{qi2017pointnet}  & 41.1 & 88.8 &  {97.3} & 69.8 & \uline{0.1} & 3.9  & 46.3 & 10.8 & 58.9 & 52.6 & 5.9  & 40.3 & 26.4 & 33.2 \\
& PointCNN \cite{li2018pointcnn}  & 57.3  & 92.3 &  {98.2} & 79.4 & {0.0} & 17.6 & 22.8 & 62.1 & 74.4 & 80.6 & 31.7  & 66.7 & 62.1 & 56.7 \\
& SegCloud \cite{tchapmi2017segcloud} & 48.9 & 90.1 & 96.1 & 69.9 & 0.0 & 18.4 & 38.4 & 23.1 & 75.9 & 70.4 & 58.4 & 40.9 & 13.0 & 41.6	\\
& Eff 3D Conv \cite{zhang2018efficient} & 51.8 & 79.8 & 93.9 & 69.0 & 0.2 & 28.3 & 38.5 & 48.3 & 71.1 & 73.6 & 48.7 & 59.2 & 29.3 & 33.1	\\
& TangentConv \cite{tangentconv} & 52.6 & 90.5 & 97.7 & 74.0 & 0.0 & 20.7 & 39.0 & 31.3 & 69.4 & 77.5 & 38.5 & 57.3 & 48.8 & 39.8	\\
& RNN Fusion \cite{ye20183d} & 57.3 & 92.3 & 98.2 & 79.4 & 0.0 & 17.6 & 22.8 & 62.1 & 74.4 & 80.6 & 31.7 & 66.7 & 62.1 & 56.7	\\
& SPGraph \cite{landrieu2018large} & 58.0 & 89.4 & 96.9 & 78.1 & 0.0 & 42.8 & 48.9 & 61.6 & 84.7 & 75.4 & 69.8 & 52.6 & 2.1 & 52.2	\\
& ParamConv \cite{wang2018deep} & 58.3 & 92.3 & 96.2 & 75.9 & 0.3 & 6.0 & 69.5 & 63.5 & 66.9 & 65.6 & 47.3 & 68.9 & 59.1 & 46.2	\\
& PointWeb \cite{zhao2019pointweb}  & 60.3  & 92.0 &  {98.5} & 79.4 & {0.0} & 21.1 & 59.7 & 34.8 & 76.3 & 88.3 & 46.9  & 69.3 & 64.9 & 52.5 \\
& RandLA-Net \cite{hu2019randla} & 63.0 & 92.4 & 96.7 & 80.6 & 0.0 & {18.3} & \uline{61.3} & 43.3 & 77.2 & 85.2 & \uline{71.5} & 71.0 & {69.2} & 52.3 \\ 
& KPConv rigid \cite{thomas2019kpconv} & \uline{65.4} & \uline{ 92.6} & {97.3} & {81.4} & 0.0 & 16.5 & 54.5 & \uline{69.5} & {80.2} & \uline{90.1} & 66.4 & {74.6} & 63.7 & {58.1}    \\
& MinkowskiNet \cite{Choy20194DSC} & \uline{65.4} & 91.8 & \uline{98.7} & 86.2 & 0.0 & \uline{34.1} & {48.9} & 62.4 & \uline{81.6} & 89.8 & {47.2} & \uline{74.9} & \uline{ 74.4} & \uline{58.6}          \\ 
\hline
%\begin{tabular}[c]{@{}l@{}}Weak \\supervision\\ (10\%)\end{tabular}     
Sparse Point Annotations (10\%) & Xu \cite{xu2020weakly}          & 48.0          & \textbf{90.9} & \textbf{97.3} & \textbf{74.8} & 0.0       & 8.4           & 49.3          & 27.3          & \textbf{71.7} & {69.0} & 53.2          & 16.5          & \textbf{63.7} & \textbf{58.1} \\
%\begin{tabular}[c]{@{}l@{}}Weak \\supervision\\ (1pt)\end{tabular} 
Sparse Point Annotations (1pt) & Xu \cite{xu2020weakly}         & 44.5          & 90.1          & 97.1          & 71.9          & 0.0       & 1.9           & 47.2          & 29.3          & 64.0          & 62.9          & 42.2          & 15.9          & 18.9          & 37.5          \\ \hline
Box + Subcloud-level Tags & \textbf{3D GrabCut}         & 57.7 & 80.3          & 88.7          & 69.6          & 0.0       & \textbf{28.8} & 61.0 & 35.2 & 66.5          & 71.7          & 69.2 & \textbf{69.7} & 61.7          & 48.2          \\ 
Box + Subcloud-level Tags & \textbf{\nickname{} (AST)}         & \textbf{60.4} & 82.0          & 92.0          & 70.8          & 0.0       & \text{28.8} & \textbf{61.9} & \textbf{38.1} & 71.4          & \textbf{85.3}          & \textbf{74.3} & 68.4 &  63.6          & 48.0          \\

\Xhline{3.0\arrayrulewidth}
\end{tabular}}

\caption{\gary{Comparative segmentation performance of difference methods on the S3DIS dataset (\textit{Area-5} split). Note that, we show the best results of fully and weakly supervised learning methods in underline and boldface, respectively. }}\label{tab:The-performance-S3DIS}.
\end{table*}
\subsection{Implementation Details}

\new{The proposed \nickname{} takes RandLA-Net \cite{hu2019randla}, an efficient framework for large-scale point clouds, as the backbone to learn point-wise semantic labels from box-level supervision and sub-cloud tags. However, our \nickname{} framework is flexible and allows using different backbone networks such as \cite{e3d, qi2017pointnet++}. During background pseudo label generation, a 1$\times$1 convolution layer is appended after the encoder of RandLA-Net to build the classifier. Once the pseudo labels of foreground and background points are generated, we train our \nickname{} using the pseudo labels with up to 100 epochs. Adam \cite{kingma2014adam} is used as the optimizer, the learning rate is set to 0.01 and decreased by 5\% after each epoch. $\alpha$ is set to 0.001 in our framework. Note that, we only utilize the spatial coordinates (\textit{i.e.,} $xyz$) of points for the generation of pseudo labels and the training of our segmentation model. All experiments are conducted on a PC with an NVIDIA GTX 1080Ti GPU with 11G memory for a fair comparison.}

%\gary{The proposed \nickname{} takes RandLA-Net \cite{hu2019randla}, an efficient framework for large-scale point clouds, as the backbone to learn point-wise semantic labels. However, our \nickname{} framework is quite flexible and allows to use any point-based method as its backbone. Note that, we append one 1$\times$1 convolution layer after the encoder of RandLA-Net to build the classifier when generating pseudo labels for background points. Once we obtain the pseudo labels, we train our \nickname{} with up to 100 epochs. Adam is used as the optimizer, the learning rate is set to 0.01 and decreased by 5\% after each epoch. Mean Intersection over Union (mIoU) is used as the main evaluation metrics. We set $\alpha$ to 0.001 in our framework. All experiments are conducted on a PC with an NVIDIA GTX 1080Ti GPU with 11G memory for a fair comparison.} \todo{In all experiments, we only use $xyz$ values of points for the generation of pseudo labels and the training of our segmentation model.}

\textbf{Annotation Cost of Box-Level Supervision.} \new{Different from existing fully-supervised segmentation pipelines \cite{hu2019randla, thomas2019kpconv, qi2017pointnet, qi2017pointnet++, li2018pointcnn, wu2018pointconv}, where dense point-wise semantic labels are used by default, the proposed \nickname{} learn semantics from low-cost bounding box annotations and subcloud-level tags. Note that, it is usually much easier to obtain 3D bounding box annotations and subcloud-level tags than dense point-wise annotations. For example, the average number of instances in a scene in the ScanNet dataset \cite{Dai2017scannet} is around 20, and the annotation cost is around 3 minutes\footnote{It took around 7s to annotate an instance with bounding boxes \cite{caesar2020nuscenes}.} using off-the-shelf labeling tools \cite{zimmer20193d, caesar2020nuscenes}. In contrast, it takes 22.3 minutes to conduct point-wise annotation \cite{Dai2017scannet} per scene. Moreover, bounding box annotations are even ready-to-use in several 3D object detection datasets \cite{Waymo, caesar2020nuscenes}.}

\section{Experiments}
\new{To validate the effectiveness of the proposed \nickname{} framework, we conduct extensive experiments on two public benchmarks, including S3DIS \cite{2D-3D-S} and ScanNet \cite{Dai2017scannet} in this section. Further, a number of ablation studies are also performed to verify the universality and robustness of the \nickname{} framework.}

\subsection{Evaluation on the S3DIS Dataset}

\begin{figure*}[tbh]
\centering
\includegraphics[width=0.9\linewidth]{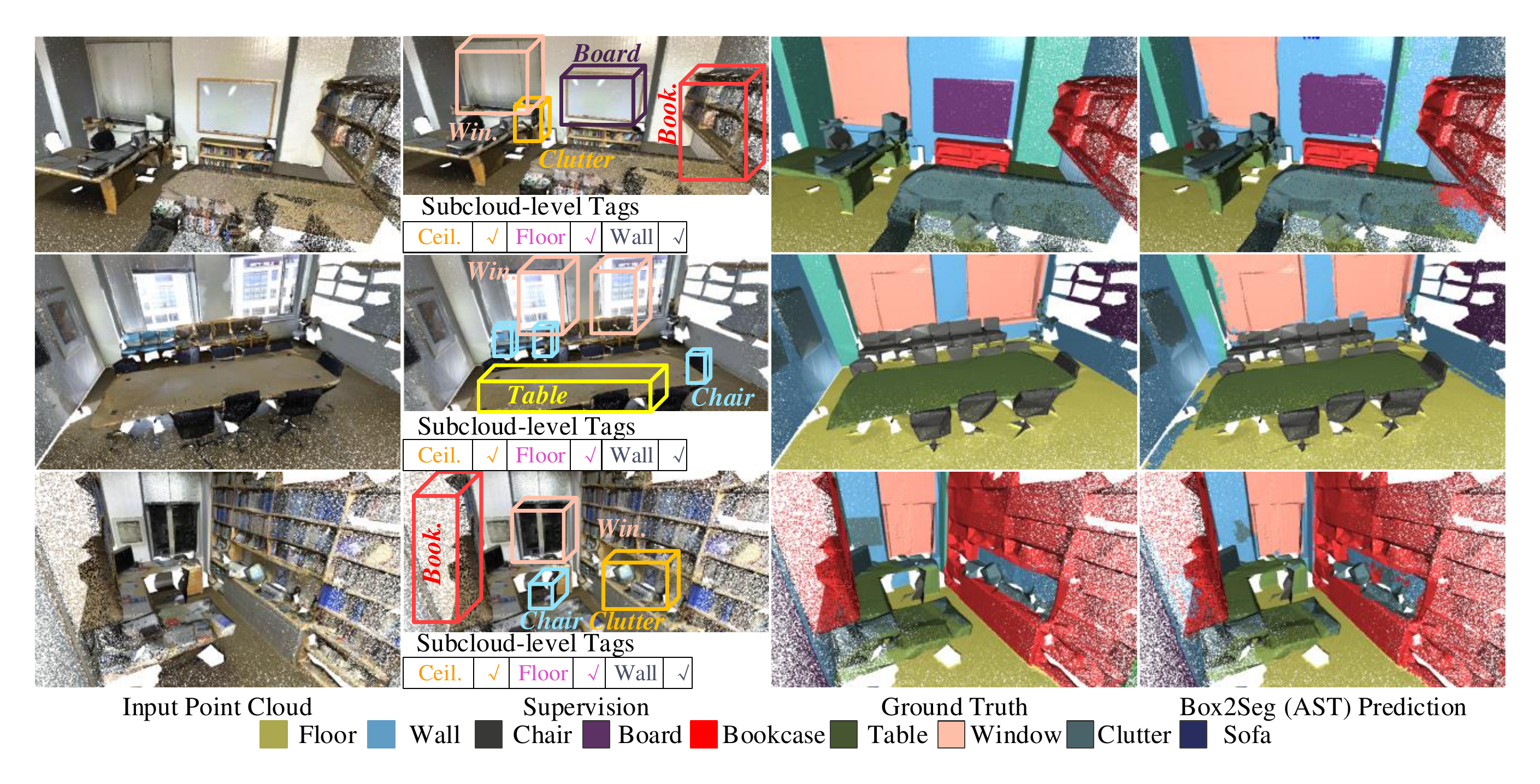}
\caption{ Visualization of the semantic segmentation results of the proposed Box2Seg (AST) on the S3DIS dataset.\label{fig:vis-s3dis}}
\end{figure*}

\begin{table*}[thb]
\centering
\resizebox{\textwidth}{!}{%
\begin{tabular}{c|rccccccccccccccccccccc}
\Xhline{3.0\arrayrulewidth}
Settings & Method & \textbf{mIoU(\%)} & \textit{bath} & \textit{bed} & \textit{bkshf} & \textit{cab} & \textit{chair} & \textit{cntr} & \textit{curt} & \textit{desk} & \textit{door} & \textit{floor} & \textit{other} & \textit{pic} & \textit{fridg} & \textit{show} & \textit{sink} & \textit{sofa} & \textit{table} & \textit{toil} & \textit{wall} & \textit{wind} \\ \hline
\multirow{15}{*}{\begin{tabular}[c]{@{}c@{}}Full \\ supervision\end{tabular}}
 & ScanNet \cite{Dai2017scannet} & 30.6 & 20.3 & 36.6 & 50.1 & 31.1 & 52.4 & 21.1 & 0.2 & 34.2 & 18.9 & 78.6 & 14.5 & 10.2 & 24.5 & 15.2 & 31.8 & 34.8 & 30.0 & 46.0 & 43.7 & 18.2 \\
 & PointNet++ \cite{qi2017pointnet++} & 33.9 & 58.4 & 47.8 & 45.8 & 25.6 & 36.0 & 25.0 & 24.7 & 27.8 & 26.1 & 67.7 & 18.3 & 11.7 & 21.2 & 14.5 & 36.4 & 34.6 & 23.2 & 54.8 & 52.3 & 25.2 \\
 & SPLATNET3D \cite{su2018splatnet} & 39.3 & 47.2 & 51.1 & 60.6 & 31.1 & 65.6 & 24.5 & 40.5 & 32.8 & 19.7 & 92.7 & 22.7 & 0.0 & 0.1 & 24.9 & 27.1 & 51.0 & 38.3 & 59.3 & 69.9 & 26.7 \\
 & Tangent-Conv \cite{tangentconv} & 43.8 & 43.7 & 64.6 & 47.4 & 36.9 & 64.5 & 35.3 & 25.8 & 28.2 & 27.9 & 91.8 & 29.8 & 14.7 & 28.3 & 29.4 & 48.7 & 56.2 & 42.7 & 61.9 & 63.3 & 35.2 \\
 & PointCNN \cite{li2018pointcnn} & 45.8 & 57.7 & 61.1 & 35.6 & 32.1 & 71.5 & 29.9 & 37.6 & 32.8 & 31.9 & 94.4 & 28.5 & 16.4 & 21.6 & 22.9 & 48.4 & 54.5 & 45.6 & 75.5 & 70.9 & 47.5 \\
 & PointConv \cite{wu2018pointconv} & 55.6 & 63.6 & 64.0 & 57.4 & 47.2 & 73.9 & 43.0 & 43.3 & 41.8 & 44.5 & 94.4 & 37.2 & 18.5 & 46.4 & 57.5 & 54.0 & 63.9 & 50.5 & 82.7 & 76.2 & 51.5 \\
 & SPH3D-GCN \cite{SPH3D} & 61.0 & 85.8 & 77.2 & 48.9 & 53.2 & 79.2 & 40.4 & 64.3 & 57.0 & 50.7 & 93.5 & 41.4 & 4.6 & 51.0 & 70.2 & 60.2 & 70.5 & 54.9 & 85.9 & 77.3 & 53.4 \\
 & KPConv \cite{thomas2019kpconv} & 68.4 & 84.7 & 75.8 & 78.4 & 64.7 & 81.4 & 47.3 & 77.2 & 60.5 & 59.4 & 93.5 & 45.0 & 18.1 & 58.7 & 80.5 & 69.0 & 78.5 & 61.4 & 88.2 & 81.9 & 63.2 \\
 & SparseConvNet \cite{sparse} & 72.5 & 64.7 & 82.1 & 84.6 & 72.1 & 86.9 & 53.3 & 75.4 & 60.3 & 61.4 & 95.5 & 57.2 & 32.5 & 71.0 & 87.0 & 72.4 & 82.3 & 62.8 & 93.4 & 86.5 & 68.3 \\
 & SegGCN \cite{lei2020seggcn} & 58.9 & 83.3 & 73.1 & 53.9 & 51.4 & 78.9 & 44.8 & 46.7 & 57.3 & 48.4 & 93.6 & 39.6 & 6.1 & 50.1 & 50.7 & 59.4 & 70.0 & 56.3 & 87.4 & 77.1 & 49.3 \\
 & RandLA-Net \cite{hu2019randla} & 64.5 & 77.8 & 73.1 & 69.9 & 57.7 & 82.9 & 44.6 & 73.6 & 47.7 & 52.3 & 94.5 & 45.4 & 26.9 & 48.4 & 74.9 & 61.8 & 73.8 & 59.9 & 82.7 & 79.2 & 62.1 \\ 
 & MinkowskiNet \cite{4dMinkpwski} & 73.6 & {\ul 85.9} & 81.8 & 83.2 & 70.9 & 84.0 & 52.1 & 85.3 & 66.0 & 64.3 & 95.1 & 54.4 & 28.6 & 73.1 & {\ul 89.3} & 67.5 & 77.2 & 68.3 & 87.4 & 85.2 & 72.7 \\
 & Occuseg \cite{han2020occuseg} & {\ul 76.4} & 75.8 & 79.6 & 83.9 & {\ul 74.6} & {\ul 90.7} & {\ul 56.2} & 85.0 & 68.0 & {\ul 67.2} & {\ul 97.8} & {\ul 61.0} & {\ul 33.5} & {\ul 77.7} & 81.9 & {\ul 84.7} & {\ul 83.0} & 69.1 & {\ul 97.2} & 88.5 & 72.7 \\
 & Virtual MVFusion \cite{kundu2020virtual} & 74.6 & 77.1 & {\ul 81.9} & {\ul 84.8} & 70.2 & 86.5 & 39.7 & {\ul 89.9} & {\ul 69.9} & 66.4 & 94.8 & 58.8 & 33.0 & 74.6 & 85.1 & 76.4 & 79.6 & {\ul 70.4} & 93.5 & {\ul 86.6} & {\ul 72.8} \\ \hline
\multirow{4}{*}{\begin{tabular}[c]{@{}c@{}}Weak\\ supervision\end{tabular}}
 & WyPR $^\dagger$ \cite{spatial_layout}  & 31.1 & 32.3 & 54.3 & 33.3 & 6.6 & 35.2 & 26.6 & 69.4 & 27.9 & 9.3 & 77.1 & 8.7 & 4.8 & 8.1 & 27.9 & 25.4 & 40.9 & 29.6 & 24.1 & 52.0 & 28.7\\
 & MPRM \cite{mprm} & 47.4 & 48.7 & \textbf{71.8} & \textbf{73.8} & 33.2 & 50.4 & 21.5 & \textbf{72.0} & 49.5 & \textbf{42.1} & 57.3 & 34.4 & 28.0 & 38.8 & 44.1 & 20.0 & 69.8 & 47.9 & 42,4 &58.0 &44.9\\
 & \textbf{ 3D Grabcut} & \textbf{56.9} & \textbf{67.6} & 69.6 & 65.7 & \textbf{49.7} & \textbf{77.9} & \textbf{42.4} & 54.8 & 51.5 & 37.6 & 90.2 & 42.2 & \textbf{35.7} & 37.9 & \textbf{45.6} & \textbf{59.6} & 65.9 & 54.4 & \textbf{68.5} & \textbf{66.5} & 55.6 \\
 
 & \textbf{\nickname{} (AST)} & 56.6 & 62.6 & 68.6 & 64.9 & 47.6 & 77.2 & 39.0 & 56.1 & \textbf{54.9} & 35.5 & \textbf{90.7} & \textbf{42.3} & 35.3 & \textbf{39.6} & 44.2 & 55.0 & \textbf{71.0} & \textbf{57.9} & 66.8 & 65.6 & \textbf{56.6} \\
 
 \Xhline{3.0\arrayrulewidth}
\end{tabular}%
}
\caption{\gary{Quantitative results of different approaches on ScanNet (online test set). Mean IoU (mIoU, \%) and per-class IoU (\%) scores are reported. $^\dagger$The results on the test set is not available yet, hence we report the result on the validation set here.}}
\label{tab:scannet_full}
\end{table*}

\new{The S3DIS \cite{2D-3D-S} dataset is composed of 272 scenes collected from 6 areas, with 273 million points being divided into 13 semantic categories, including \textit{ceiling}, \textit{floor}, \textit{tables}, \textit{chairs}, \textit{etc}. Following \cite{thomas2019kpconv, pointweb}, \textit{Area-5} is selected as the validation set to evaluate the final segmentation performance of our \nickname{}.  Mean Intersection over Union (mIoU) is used as the main evaluation metrics. Note that, the box-level labels in this dataset are not orginally provided. Therefore, we first generate box-level labels and subcloud-level tags based on the semantic and instance labels provided in the dataset. Three categories, including \textit{ceiling},  \textit{floor}, and \textit{wall}, are considered as the background categories, while the rest are considered as foreground classes.} 

\new{The quantitative performance comparison between our \nickname{} and other methods is shown in Table \ref{tab:The-performance-S3DIS}. It can be seen that the proposed \nickname{} can achieve better performance than existing weakly supervised approaches \cite{xu2020weakly}, especially in light of the bounding box annotations and subcloud-level tags are much easier and cheaper than sparse point-level annotations (\textit{e.g.} annotating 10\% or 1\% points). Moreover, there is no need to develop any customized labeling tools when adopting our weak supervision scheme, several off-the-shelf tools can be seamlessly integrated into our pipeline. We also noticed      \nickname{} with the proposed attention-based self-training pipeline can achieve better performance than the 3DGrabcut based baseline. Overall, there is still a certain performance gap between the proposed method and its fully-supervised counterpart (\textit{i.e.}, RandLA-Net \cite{hu2019randla}), primarily because the supervision for background points (\textit{i.e.}, sub-cloud tags) are too weak for the network to learn a meaningful representation. This can be also reflected from the relatively poor performance on the background categories (about 8\% decrease in \textit{ceiling}, \textit{floor}, and \textit{wall} in average).}

\new{We also show the visualization results on the S3DIS dataset in  Fig. \ref{fig:vis-s3dis}. It can be seen that the overall segmentation performance of the proposed \nickname{} is satisfactory except for the few misclassifications at the boundary areas. In particular, our \nickname{} only takes the box-level supervision and subcloud-level tags as its supervision signals.}

\subsection{Evaluation on the ScanNet Dataset}
%ScanNet \cite{Dai2017scannet} are consisted by 1613 indoor scenes. We use Scannet V2 in our experiments, which includes 20 categories totally. 1201 scenes and 312 scenes are used for training and validation respectively, and 100 scenes are used for online testing. In this dataset, floor and wall are background classes.

\new{The ScanNet \cite{Dai2017scannet} dataset consists of 1613 3D scenes, with 1201 scenes, 312 scenes, and 100 scenes being used for training, validation, and online test, respectively. Similar to the S3DIS dataset,  the bounding box annotations are generated from the provided instance labels. In this dataset, only \textit{floor} and \textit{wall} are considered as background classes. Note that, a number of points are denoted as unclassified in this dataset.
}

\new{It can be seen that our \nickname{} outperforms existing weak supervision frameworks MPRM \cite{mprm} and WyPR \cite{spatial_layout}, which use subcloud-level labels and scene-level tags, respectively. In addition, our method also achieves better performance than several fully supervised methods such as PointNet++ \cite{qi2017pointnet++} and PointCNN \cite{li2018pointcnn}, with the usage of weak supervision signals only. }

\subsection{Ablation Study}

\revise{
\textbf{\textit{-Can \nickname{} trained with noisy bounding box annotations?}} \new{Considering that the bounding box annotations in practice are usually not accurate as the ones we generate directly from the point-wise instance labels, thereby we further verify the robustness of the proposed method in this section. In particular, we apply three different perturbations on the ground-truth bounding boxes, including random translation/scaling/discarding, to simulate the noisy bounding box annotations in real situations. Specifically, random translation is achieved by shifting the center of a bounding box, the amount of shift is randomly generated in a specific range (e.g. [-10$\%$, 10$\%$], [-20$\%$, 20$\%$]) of the bounding box size. Similarly, random scaling is achieved by changing the scale of a bounding box. The scale ratios are also randomly generated within the range of [0.9, 1.1]. Finally, random discarding is achieved by randomly dropping a fraction of bounding box annotations. The drop ratio is varied from 0 to 20$\%$. We then train our method with the noisy bounding boxes. The experimental results are shown in Table \ref{tab:The-performance-with-noise}. We can see that the proposed \nickname{} can achieve relatively robust segmentation performance (±10$\%$), even only provided with the inaccurate bounding box supervision, further verifying the robustness of the proposed framework. Please refer to the appendix for detailed results.}}

\begin{table}[tbh]
\centering
\resizebox{0.4\textwidth}{!}{
\begin{tabular}{lcc}
\Xhline{2.0\arrayrulewidth}
{ Methods} & mIoU(\%) \\ \hline
\Xhline{1.0\arrayrulewidth}
{ BBox with random translation (±10$\%$)}  & 56.8\\
{ BBox with random translation (±20$\%)$} & 53.6\\
{ BBox with random scaling (scale ratio: [0.9~1.1])}& 56.4 \\
{ BBox with random scaling (scale ratio: [0.8~1.2])}  & 55.0\\
{ BBox with random discarding (10$\%$)} & 57.1\\
{ BBox with random discarding (20$\%$)}& 53.9\\
{ \textbf{Ground-truth BBox}} & \textbf{60.4}\\
\Xhline{2.0\arrayrulewidth}
\end{tabular}}
\caption{\new{Quantitative results achieved by our \nickname{} on Area-5 of the S3DIS \cite{2D-3D-S} dataset with noisy bounding box annotations.
}}
\label{tab:The-performance-with-noise}
\end{table}

\textbf{\textit{-What is the impact of each component?}} Our \nickname{} has several major components, including pseudo labeling, attention mechanism, and pseudo labels refinement. The ablative results achieved on the S3DIS dataset with different combinations of these components are shown in Table  \ref{tab:The-performance-each-element}. 
\new{It can be seen that: 1) The refinement step has the greatest positive impact on the overall segmentation performance. This is primarily because this step can largely filter most of the unreliable pseudo labels (i.e., predictions), avoiding the erroneous gradients back propagated throughout the network. 2) The incorporation of attention and pseudo labeling can further improve the overall performance, but not significantly as that of the refinement step.}

\begin{table}[tbh]
\centering
\resizebox{0.4\textwidth}{!}{
\begin{tabular}{cccc}
\Xhline{2.0\arrayrulewidth}
Refinement & Attention Map Modulation & Pseudo Labeling & mIoU($\%$) \\
\Xhline{2.0\arrayrulewidth}
 &  &  & 54.4 \\
 
 $\surd$&  &  & 59.5 \\
  $\surd$ & $\surd$ &  &  60.1\\
$\surd$ & $\surd$ & $\surd$ & \textbf{60.4}\\
\Xhline{2.0\arrayrulewidth}
\end{tabular}}
\caption{\new{Quantitative results achieved by our \nickname{}  on Area-5 of the S3DIS \cite{2D-3D-S} dataset with ablative components.}}
\label{tab:The-performance-each-element}
\end{table}

% \textbf{\textit{-Can \nickname{} use different backbones?}} \gary{To evaluate the versatility of our framework, we further develop a \nickname{} variant by adopting PointNet++ \cite{qi2017pointnet++} as its backbone. }

\begin{comment}
As the \nickname{} can achieve good semantic segmentation performance on the backbone based on the RandLA-Net. One may wonder that the proposed method is still effective and useful if we employ an alternative point cloud segmentation module. To verify whether the proposed method can be applied to other frameworks, we conduct an ablation study that adopts PointNet++ \cite{qi2017pointnet++} as the backbone. 
\end{comment}

\textbf{\textit{-Can \nickname{} trained with different backbones?}} \new{To evaluate the versatility of our framework, we further develop a \nickname{} variant by adopting PointNet++ \cite{qi2017pointnet++} as its backbone. The results achieved by ablated networks on the S3DIS dataset \cite{2D-3D-S} are shown in Table \ref{tab:The-performance-S3DIS-PointNet++}. It can be seen that  \nickname{} (PointNet++) can also achieve a reasonable segmentation performance, although the performance gap with the fully supervised method has become larger. This is likely because of the decline in feature extraction capabilities. }

\begin{table}[tbh]
\centering
\resizebox{0.4\textwidth}{!}{%
\begin{tabular}{crc}
\Xhline{3.0\arrayrulewidth}
Settings & Method & mIoU($\%$) \\ 
\Xhline{2.0\arrayrulewidth}
\multirow{2}{*}{Full supervision}
 & PointNet++ \cite{qi2017pointnet++}& 52.3 \\
 & RandLA-Net \cite{hu2019randla}& \uline{63.0} \\ 
 \Xhline{2.0\arrayrulewidth}
\multirow{2}{*}{Weak supervision}  & \textbf{Box2Seg} (PointNet++) & 44.5 \\
& \textbf{Box2Seg} (RandLA-Net) & \textbf{60.4} \\
\Xhline{3.0\arrayrulewidth}
\end{tabular}}
\caption{\gary{ Quantitative results achieved by \nickname{} with different backbone networks on the S3DIS \cite{2D-3D-S} dataset.}}
\label{tab:The-performance-S3DIS-PointNet++}
\end{table}

% \ly{
% \textbf{\textit{-Can \nickname{} trained without points in overlapping region?}} In our experiments, we adopt the self-training framework for pseudo labels assignment of points located in the overlapping region. To verify the effectiveness of the training strategy, we conduct an experiment ignore all ambiguous points, specifically, the gradients of these points are not allowed flowing into the training pipeline. Experimental results are shown in Tab. \ref{tab:The-performance-with-no-self-training}. As seen, leveraging points located in the overlap region can boost the segmentation performance of the  \nickname{} framework. Besides, compared with the unsupervised 3DGrabcut framework, the training based method is easy to implement and demonstrates the effectiveness of using self-training based approach to capture semantic information of unlabeled raw point cloud.
% }

\begin{comment}
\section{Discussions}
\qy{Although good performance has achieved by the proposed \nickname{}, it remains several questions that need to be further investigated. First, the proposed framework is not end-to-end trainable. Second, the bounding box labels used in our experiment are relatively ideal (\textit{i.e.}, generate from the instance labels). Future work will further explore the case of taking inaccurate bounding box annotations as supervision signals.}
\end{comment}

\section{Conclusion}
%\gary{In this paper, we propose a weakly supervised framework for semantic segmentation of unstructured 3D point clouds. The key of our approach is to learn pseudo labels inside and outside bounding boxes through a divide-and-conquer manner. Besides, we explore an unsupervised and a self-training-based method to assign pseudo labels to points inside bounding boxes. Our experiments show that satisfactory semantic segmentation results can be achieved with only bounding box annotations and subcloud-level tags. It would also be interesting to further generalize our framework to 3D instance segmentation and panoptic segmentation in future.}

\new{In this paper, we propose a weakly-supervised framework for semantic segmentation of unstructured 3D point clouds. The key of our approach is to learn pseudo labels inside and outside bounding boxes through a divide-and-conquer manner. Specifically, an attention-based self-training pipeline and point class activation mapping techniques are introduced to generate foreground and background pseudo labels. Extensive experiments on two datasets demonstrate the effectiveness of the proposed framework. However, we also notice that the performance on background points is still limited due to insufficient supervision. Future works will further explore self-supervised pretraining or unsupervised semantic clustering.}

{\small
\bibliographystyle{ieee_fullname}
\bibliography{egbib}
}

\end{document}